\pdfoutput=1

\documentclass[11pt]{article}

\usepackage[final]{acl}

\usepackage{times}
\usepackage{latexsym}

\usepackage[T1]{fontenc}

\usepackage[utf8]{inputenc}

\usepackage{microtype}

\usepackage{inconsolata}

\usepackage{multirow}
\usepackage{tabularx}
\usepackage{float}
\usepackage{placeins}
\usepackage{booktabs}
\usepackage{array}
\usepackage{graphicx}

\usepackage[T1]{fontenc}
\usepackage[american,german]{babel}
\usepackage{caption}

\title{OPDAI at SemEval-2024 Task 6: Small LLMs can Accelerate Hallucination Detection with Weakly Supervised Data}

\author{Chengcheng Wei,  Ze Chen,  Songtan Fang, Jiarong He \and Max Gao\\
        Interactive Entertainment Group of Netease Inc., Guangzhou, China\\
        \{weichengcheng, jackchen, fangsongtan, gzhejiarong, jgao\}@corp.netease.com}

\begin{document}
\maketitle
\begin{abstract}
This paper mainly describes a unified system for hallucination detection of LLMs, which wins the second prize in the model-agnostic track of the SemEval-2024 Task 6, and also achieves considerable results in the model-aware track. This task aims to detect hallucination with LLMs for three different text-generation tasks without labeled training data. We utilize prompt engineering and few-shot learning to verify the performance of different LLMs on the validation data. Then we select the LLMs with better performance to generate high-quality weakly supervised training data, which not only satisfies the consistency of different LLMs, but also satisfies the consistency of the optimal LLM with different sampling parameters. Furthermore, we finetune different LLMs by using the constructed training data, and finding that a relatively small LLM can achieve a competitive level of performance in hallucination detection, when compared to the large LLMs and the prompt-based approaches using GPT-4.
\end{abstract}

\section{Introduction}

The emergence of Large Language Models (LLMs)\cite{zhao2023survey} has sparked a significant transformation in the field of Natural Language Processing (NLP), ushering in a new era of unparalleled advancements in text generation and comprehension. This revolutionary technology has elevated the capabilities of AI systems, enabling them to perform complex reasoning and problem-solving tasks with remarkable proficiency\cite{zhao2023survey}. At the heart of this transformation lies the LLMs' ability to compress vast amounts of knowledge into neural networks, effectively turning them into versatile agents capable of engaging in natural language conversations with humans\cite{hadi2023survey}. This has broadened the scope of AI applications beyond traditional domains such as chatbots and virtual assistants, into areas previously thought to be the exclusive domain of humans, particularly those involving creativity and expertise. LLMs are not only limited to language-related tasks but can also function as generalist agents, collaborating with external systems, tools, and models to achieve a wide range of objectives set by humans\cite{triguero2024general}.

However, recent advancements in research have uncovered a concerning weakness: their proneness to hallucinate content across a range of applications\cite{ ji2023survey}. Hallucination is defined as the generation of information that either conflicts with established sources or cannot be substantiated by available knowledge. The occurrence of hallucination in LLMs poses a significant threat to their practical deployment. While prior works have delved into the roots of hallucination within specific, smaller-scale language models and tasks, there is still a notable gap in understanding the exact nature and prevalence of content that LLMs are likely to hallucinate\cite{cui2024risk, chang2023survey}.

To address this challenge, we implements a unified system for hallucination detection of LLMs, when there is no labeled training data. This system comprises five parts: \textit{Base Model Selection}, \textit{Prompt Engineering}, \textit{Weakly-supervised Data Generation}, \textit{SFT} and \textit{Ensemble Learning}. We first verify the performance of different base LLMs on this task. Then we select the best LLMs and prompt is optimized to improve the performance. And weakly-supervised dataset is generated by using the selected LLMs. For further improvement, SFT is done based on the constructed dataset and ensemble learning is adopted.

\section{Task Description}

\begin{table}[h]
\centering
\begin{tabular}{c|c|c|c}
\hline
\textbf{\begin{tabular}[c]{@{}c@{}}trial data\end{tabular}} & \textbf{\begin{tabular}[c]{@{}c@{}}unlabled \\ train data\end{tabular}} & \textbf{\begin{tabular}[c]{@{}c@{}}validation\\  data\end{tabular}} & \textbf{\begin{tabular}[c]{@{}c@{}}test \\ data\end{tabular}} \\ \hline
80                                                              & 60000                                                                   & 1000                                                                & 3000                                                          \\ \hline
\end{tabular}
\caption{Dataset provided by SHROOM}
\label{tab:shroom} 
\end{table}

SHROOM asked participants to perform binary classification to identify cases of fluent overgeneration hallucinations in two different setups: model-aware and model-agnostic tracks. And three different NLG tasks: definition modeling (DM), machine translation (MT) and paraphrase generation (PG) are covered in both tracks. In model-aware track, the model information is provided. The provided development and test sets include binary annotations from a minimum of five different annotators, along with a majority vote gold label. Table \ref{tab:shroom} gives an overview of the provided dataset.

\section{Methodology}
\label{tab:method}

\begin{figure*}[]
  \centering
  \includegraphics[width=1.0\textwidth]{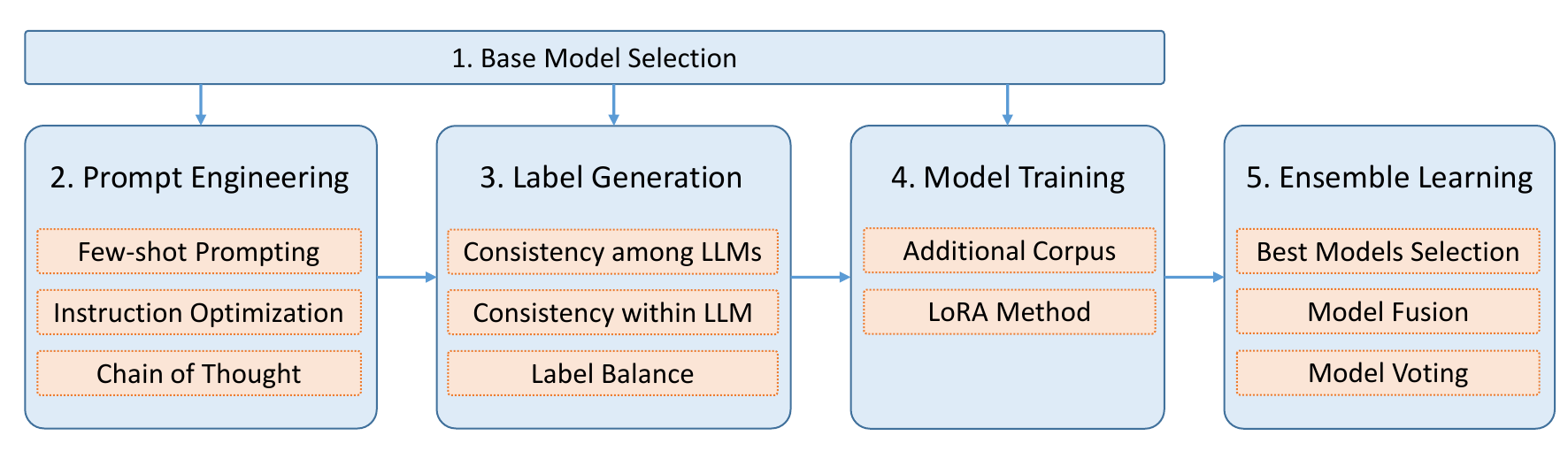}
  \caption{Overview of our proposed method. The method consists of five main steps, each of which comprises several modules that we have designed for SemEval-2024 Task 6. }
  \label{fig:overview}
\end{figure*}

Figure~\ref{fig:overview} shows the overview of our approach. Our method consists of five main steps. First of all, multiple LLMs are compared on the hallucination detection validation dataset and among which we select the best base model. The LLMs selected in the first step will be utilized in the subsequent steps 2, 3, and 4. In the second step, we designed a Prompt Engineering module consisting of three sub-modules: few-shot prompting, instruction optimization, and the utilization of Chain-of-Thought. The subsequent experiments will demonstrate that prompt engineering significantly enhances the capabilities of the base model.

Then moving on to the Label Generation step, we apply the Prompt Engineering module to the selected best LLMs. We make predictions on the unlabeled training set and ensure the inference consistency among multiple LLMs as well as the inference consistency under different inference parameters of individual LLM. We also ensure label balance in this process.

The following steps are Model Training and Ensemble Learning. We utilize SFT based on the constructed dataset using the LoRA~\cite{hu2021lora} method. Finally, we select a few top-performing models and perform fusion at both the model weight level(model fusion) and the model prediction probability level(model voting) in order to seek better performance.



\subsection{Prompt Engineering}
\label{tab:prompt_engineering}

After experimenting with multiple large-scale language models, we select the best-performing 14B \textit{Mixtral\_7Bx2\_MoE} as the base model for prompt engineering to achieve better results. 

\textit{Few-shot prompting}. Few-shot prompting can be used as a technique to enable in-context learning where we provide demonstrations from the provided \textit{trial data} in the prompt to steer the base model to better performance. Considering that the trial data contains with labels, we randomly sample the specific task's datapoints for different tasks, ensuring an equal number of data points for "hallucination" and "not hallucination" examples, to serve as few-shot examples for each respective task.

\textit{Optimizing the instruction}. There are some deficiencies with the instruction introduced in Section~\ref{tab:section_baseline}, and we name this version of instruction as the \textit{naive} version. The most obvious issue is that the naive instruction does not include the descriptions of the DM, MT, and PG tasks. The desired task description includes the task definition and all known useful information, rather than just focusing on the sentence and context. We design different instructions for different tasks, which can be found in the Appendix~\ref{sec:appendix}. In this way, we can append additional information to the prompt to assist the LLM in better understanding the problem.

\textit{Chain-of-thought prompting}. Chain-of-thought (CoT) prompting\cite{wei2022chain} is a recently developed method that encourages the language model to explain its reasoning. We combine the aforementioned few-shot prompting, developed instruction, and CoT to utilize them together to further enhance the capability of the base model.

\subsection{Label Generation and Weakly-SFT} 
\label{tab:3_2}
After improving the performance of the base model using prompt engineering, we use the optimal settings to infer on unlabeled training data and obtain weakly supervised labels. These weakly supervised labels are then used to finetune the base model.

\textit{Inference consistency in generating labels}. During the process of inferring weakly supervised labels for the unlabeled training data, we placed a great emphasis on both the consistency of inference across different LLMs and the consistency of inference within the same LLM but with different parameter settings. To achieve this, we carefully selected several sets of top-performing base models. Leveraging the prompt engineering techniques mentioned earlier, we conducted inference on the same model using various parameter configurations. Subsequently, we handpicked the data points with consistent inferences across different parameter settings to establish the final inference results for that particular LLM. Additionally, we applied a filtering process to the inference results obtained from different LLMs, ensuring that only datapoints with consistent inferences were retained. Through these rigorous steps, we ensured that our generated weakly supervised labels for the training set exhibits robustness to both the choice of LLM base and the specific sampling parameters employed. Finally, we used sampling techniques to balance the data volume of the two categories.

\textit{Fine-tuning LLMs}. The weak supervision generated by the base model is applied to guide models of equal or smaller scale. The LLMs undergo fine-tuning using the LoRA approach, a popular and lightweight training technique that effectively reduces the number of trainable parameters. Despite this reduction, the fine-tuned models maintain comparable training results to those of the full parameter models. The best checkpoint model files are selected from the validation set during this fine-tuning process. 

\subsection{Ensemble Learning}
We also propose an ensemble learning approach for performance improvement, utilizing fusion strategies at both the model level and the inference level.

\textit{Model fusion}. MergeKit\footnote{\url{https://github.com/arcee-ai/mergekit}} is a toolkit designed for merging trained language models. We carefully selected a few high-accuracy models and utilized MergeKit to perform model fusion using the SLERP~\cite{shoemake1985animating}, TIES~\cite{yadav2023ties} and linear~\cite{wortsman2022model} methods. Traditionally, model merging often resorts to weight averaging which, although straightforward, might not always capture the intricate features of the models being merged. The SLERP technique addresses this limitation, producing a blended model with characteristics smoothly interpolated from both parent models, ensuring the resultant model captures the essence of both its parents. Meanwhile, the TIES method is proposed to resolve interference issues by resetting parameters, resolving sign conflicts, and merging only compatible parameters. TIES outperforms many existing methods across diverse settings, emphasizing the importance of addressing interference in model merging for enhanced performance and versatility.

\textit{Model Voting}. In addition to the model-level fusion, we also explored fusion at the probability level of model generation, which can be understood as a form of model voting. We selected another group of highly accurate candidate models and performed linear fusion at the probability level. Specifically, we calculate the weighted summation of the probability values on "existing hallucination" predicted by different candidate models for different tasks. By tuning the linear weight combination, we are able to determine the optimal combination of weights for each task. Finally, combining different tasks together yields the final fusion result. In this way, we implement weighted voting of models at the inference result level.

\section{Results and Analysis}

In this section, we will present a series of experiments to illustrate the effectiveness of our method.

\subsection{Baseline}
\label{tab:section_baseline}

\begin{table*}[]
\centering
\newcolumntype{C}{>{\centering\arraybackslash}X}
    \begin{tabularx}{\textwidth}{ccCCCC}
        \toprule
        \multirow{2}{*}{\textbf{Model Name}}    & \multirow{2}{*}{\textbf{Model Size}} & \multicolumn{2}{c}{\textbf{Model-agnostic track}} & \multicolumn{2}{c}{\textbf{Model-aware track}} \\
                                       &      & \textit{\textbf{acc}}  & \textit{\textbf{rho}}  & \textit{\textbf{acc}} & \textit{\textbf{rho}}      \\
        \hline
        \addlinespace
        Mistral-7B-Instruct-v0.2-GGUF  & 7B                          & 0.649               & 0.380              & 0.707             & 0.461             \\
        Mistral-7B-Instruct-v0.2       & 7B                          & 0.655               & 0.375              & 0.705             & 0.468             \\
        Mixtral\_7Bx2\_MoE             & 14B                         & \textbf{0.747}      & 0.518              & 0.764             & 0.475             \\
        Mixtral-8x7B-Instruct-v0.1     & 46.7B                       & 0.723               & 0.526              & 0.745             & 0.552             \\
        Nous-Hermes-2-Mixtral-8x7B-DPO & 46.7B                       & 0.741               & 0.607              & \textbf{0.766}    & 0.614             \\
        Nous-Hermes-2-SOLAR-10.7B      & 10.7B                       & 0.725               & 0.592              & 0.722             & 0.588             \\
        SOLAR-10.7B-Instruct-v1.0      & 10.7B                       & 0.737               & 0.438              & 0.747             & 0.381             \\
        SauerkrautLM-SOLAR-Instruct    & 10.7B                       & 0.733               & 0.418              & 0.752             & 0.368             \\
        Sakura-SOLAR-Instruct-DPO-v2   & 10.7B                       & 0.733               & 0.426              & 0.745             & 0.357             \\
        \bottomrule
        \hline
    \end{tabularx}
\caption{The performance of different-sized LLMs on the validation set. The competition includes two tracks: model-agnostic track and model-aware track. For each track, both prediction accuracy(\textit{acc}) and Spearman's Rho value(\textit{rho}) are provided.}
\label{tab:baseline}
\end{table*}

To begin with, our initial step entails presenting the basic performance of LLMs of varying sizes on the validation set. Subsequently, we will delve into an analysis of the LLMs' capabilities in detecting hallucinations in the given task.

Throughout the experiments, we ensure the generation hyperparameters remain consistent across all LLMs. Additionally, the instruction utilized for detecting hallucinations is sourced from the official \textit{participant\_kit}. This version of the instruction is referred to as the "naive instruction" and can be located in Appendix~\ref{sec:appendix} for reference.

Table~\ref{tab:baseline} illustrates our evaluation of LLMs from both \textit{Mistral} and \textit{SOLAR} families, considering varying sizes and variants, on the validation set. In general, larger models tend to yield better results. For instance, within the \textit{Mistral}-family, the accuracy and Spearman's Rho value of the 7B model are comparatively lower than those of larger models, a trend observed in both the model-agnostic and model-aware tracks. Furthermore, LLMs of the same size exhibit diverse results in hallucination detection tasks owing to distinct fine-tuning methods employed. This observation holds true in our experiments with the \textit{SOLAR}-family, emphasizing the impact of fine-tuning on performance.

There is a noteworthy observation in Table~\ref{tab:baseline}. The medium-sized 14B \textit{Mixtral\_7Bx2\_MoE} model achieves comparable accuracy to the larger-sized 46.7B models in both the model-agnostic and model-aware tracks. This suggests that the fine-tuning approach and training corpus of the 14B model are well-suited for the hallucination detection task. Furthermore, the 14B model outperforms the 46.7B model in terms of inference speed and training cost. As a result, in the subsequent section, we will further enhance the effectiveness of the 14B model through prompt engineering.

\subsection{Performance Improvement}

\begin{table}[]
    \centering
    \begin{tabular}{cccc}
        
        \toprule
        \textbf{few-shot}   & \textbf{inst.} & \textbf{agnostic\_acc} & \textbf{aware\_acc} \\
        \hline
        \addlinespace
        \multirow{2}{*}{2-shot} & naive       & 0.745            & 0.774                \\
                                & ours    & \textbf{0.770}   & \textbf{0.806}       \\
        \addlinespace
        \multirow{2}{*}{4-shot} & naive       & 0.762            & 0.772                \\
                                & ours    & \textbf{0.782}   & \textbf{0.806}       \\
        \addlinespace
        \multirow{2}{*}{6-shot} & naive       & 0.764            & 0.774                \\
                                & ours    & \textbf{0.772}   & \textbf{0.804}       \\   
        \addlinespace
        \multirow{2}{*}{8-shot} & naive       & 0.762            & 0.772                \\
                                & ours    & \textbf{0.774}   & \textbf{0.804}       \\
        \bottomrule
        \hline
    \end{tabular}
\caption{Our proposed instruction exhibits overall superior accuracy compared to the naive version on the validation set. We applied few-shot prompting in all of the aforementioned experiments. The term "inst." stands for "instruction".}
\label{tab:task_prompt}
\end{table}

In this section, our focus is on enhancing the accuracy of the 14B \textit{Mixtral\_7Bx2\_MoE} model through prompt engineering methods.

\textit{Few-shot prompting}. A few-shot prompting approach is applied by randomly selecting an equal number of positive and negative samples as demonstrations for task definition modeling (DM), machine translation (MT), and paraphrase generation (PG). In the experimental setup, we use 2, 4, 6, and 8 examples for few-shot prompting on the \textit{Mixtral\_7Bx2\_MoE} model, while keeping the generation hyperparameters consistent with the experiments in Table~\ref{tab:baseline}. The accuracy of the few-shot prompting strategy is shown with the \textit{inst.=naive} setting in Table~\ref{tab:task_prompt}, where we observe that experiments with 4, 6, and 8 shots perform better than the zero-shot baseline(acc is 0.747 in Table~\ref{tab:baseline}) in the model-agnostic track, and all the few-shot settings experiments achieve better results than the zero-shot baseline(acc is 0.764 in Table~\ref{tab:baseline}) in the model-aware track.

\textit{Optimizing the instruction}. As discussed in Section~\ref{tab:prompt_engineering}, the naive instruction provided by the competition organizers has some limitations. To overcome these limitations, we enhanced the instructions by incorporating task-specific background knowledge and multidimensional information, taking into account the unique characteristics of each task. The improved instructions, as demonstrated with the \textit{inst.=ours} setting in Table~\ref{tab:task_prompt}, yield better performance compared to using the initial naive instruction. Notably, in the 2-shot setting, both tracks exhibited an improvement of over 2 percentage points by leveraging our proposed instructions.

\textit{Chain of thought prompting}. We adopt the CoT approach, after generating reasons for the presence or absence of hallucinations in the trial data. The experimental results of CoT are presented in Table~\ref{tab:cot}, which indicates that CoT exhibits higher efficacy in the few-shot scenario when there are more demonstrations accessible.

\begin{table}[]
    \centering
    \begin{tabularx}{0.47\textwidth}{cccc}

    \toprule
    \textbf{few-shot}       & \textbf{CoT} & \textbf{agnostic\_acc} & \textbf{aware\_acc} \\
    \hline
    \addlinespace
    \multirow{2}{*}{2-shot} & w/o          & 0.770                   & \textbf{0.806}       \\
                            & with         & 0.770                   & 0.792                \\
    \addlinespace
    \multirow{2}{*}{4-shot} & w/o          & \textbf{0.782}          & \textbf{0.806}       \\
                            & with         & 0.766                   & 0.796                \\
    \addlinespace
    \multirow{2}{*}{6-shot} & w/o          & 0.772                   & 0.804                \\
                            & with         & \textbf{0.774}          & \textbf{0.806}       \\
    \addlinespace
    \multirow{2}{*}{8-shot} & w/o          & 0.774                   & 0.804                \\
                            & with         & \textbf{0.792}          & 0.804                \\
    \bottomrule
    \hline
    \end{tabularx}
\caption{CoT demonstrates an improved capability in hallucination detection when provided with a larger number of demonstrations in few-shot prompting and utilizing our proposed instruction. The results are on the validation set.}
\label{tab:cot}
\end{table}

\subsection{Weakly-supervised Fine-tuning}

As mentioned earlier, we enhance the accuracy of hallucination detection by selecting the best baseline model and incorporating additional prompt engineering techniques. Building upon this, we leverage weak supervision by labeling the unlabeled training data for training. Subsequently, the LLMs are fine-tuned using the generated labels to further augment the capability of hallucination detection.

\begin{table*}[ht]
\centering
\newcolumntype{C}{>{\centering\arraybackslash}X}
    \begin{tabularx}{\textwidth}{ccCCCC}
        \toprule
        \multirow{2}{*}{\textbf{Model Name}}    & \multirow{2}{*}{\textbf{Model Size}} & \multicolumn{2}{c}{\textbf{Model-agnostic track}} & \multicolumn{2}{c}{\textbf{Model-aware track}} \\
                                                &      & \textit{\textbf{acc}}  & \textit{\textbf{rho}}  & \textit{\textbf{acc}} & \textit{\textbf{rho}}      \\
        \hline
        \addlinespace
        Mistral-7B-Instruct-v0.2          & 7B                          & \textbf{0.806}      & 0.708              & 0.790             & 0.699             \\
        Nous-Hermes-2-SOLAR-10.7B         & 10.7B                       & 0.764               & 0.690              & 0.806             & 0.714             \\
        SOLAR-10.7B-Instruct-v1.0         & 10.7B                       & 0.772               & 0.703              & 0.810             & 0.717             \\
        Mistral-7B-Instruct-v0.2-2x7B-MoE & 12.8B                       & 0.790               & 0.725              & \textbf{0.814}    & 0.698             \\
        Mixtral\_7Bx2\_MoE                & 14B                         & 0.780               & 0.675              & 0.796             & 0.657             \\
        \textit{raw:} Mixtral\_7Bx2\_MoE  & 14B                    & 0.792               & 0.707              & 0.804             & 0.690           \\

        \bottomrule
        \hline
    \end{tabularx}
\caption{The performance of weakly-supervised fine-tuning on the validation set. Smaller LLMs perform better than the 14B supervisor. The last line indicates the performance of Mixtral\_7Bx2\_MoE without SFT.}
\label{tab:sft}
\end{table*}

\textit{Generating weak supervision for training data}. Utilizing the 14B \textit{Mixtral\_7Bx2\_MoE} model as a foundation, we incorporate few-shot prompting, our proposed instruction, and the CoT strategy to create a supervision model known as the `8-shot' setting, as mentioned in Table~\ref{tab:cot}. This approach is applied to hallucination detection across 60,000 datapoints from both the model-agnostic and model-aware tracks, ensuring a balanced distribution of categories. Additionally, we introduce multiple optimal models, as discussed in Section~\ref{tab:method}, to ensure consistent inference across multiple models and maintain inference consistency within the same model but with different inference parameters.

\textit{Fine-tuning LLMs}. The experimental results of weakly-supervised fine-tuning are presented in Table~\ref{tab:sft}, which demonstrates that smaller models can effectively learn from the weak supervision provided by the 14B model. In some cases, these smaller models even outperform the 14B model in terms of accuracy. Notably, the 14B model fails to surpass the performance of equivalently-sized supervisor even when multiple hyper-parameter settings are employed. A comparison between lora training and full-parameter training reveals that the lora-style training yields superior results. Further details can be found in Appendix \ref{sec:ap_lora_full}.

\subsection{Ensemble Learning}

In addition to fine-tuning LLMs with weak supervision labels as mentioned in the previous section, we first combine different model checkpoints by the MergeKit tool and then perform model voting strategy to enhance performance. 


\textit{Model fusion}. We implement model merging using different modes of MergeKit, \textit{i.e.}, SLERP, TIES and Linear, in our study. Taking the accuracy optimization of the model-agnostic track as an example, let's begin by selecting three highly capable candidate models. These models are fine-tuned versions of the 7B \textit{Mistral-7B-Instruct-v0.2}, and they are different checkpoints from the same training task. The detailed training setting can be found in Appendix~\ref{sec:ap_v1_v3}. By utilizing model fusion techniques, we can achieve a maximum accuracy of 0.814 with the newly merged models. For a detailed overview of the experiments on the model-agnostic track, refer to Table~\ref{tab:mergekit}.

\begin{table}[]
\centering
\begin{tabular}{cc}
\toprule
\textbf{Method} & \textbf{agnostic\_acc} \\
\hline
Mistral-7B-Instruct-v0.2-sft-v1     & 0.804                  \\
Mistral-7B-Instruct-v0.2-sft-v2     & 0.798                  \\
Mistral-7B-Instruct-v0.2-sft-v3    & 0.808                  \\
Linear-merged model  & \textbf{0.814}         \\
SLERP-merged model   & \textbf{0.814}         \\
TIES-merged model    & \textbf{0.814}        \\
\bottomrule
\hline
\end{tabular}
\caption{Model fusion results for the model-agnostic track on the validation set. The merged models outperform any individual model in terms of accuracy.}
\label{tab:mergekit}
\end{table}

\textit{Model Voting}. We also validate the effectiveness of weighted voting at the inference result level. We select the top-performing models from the weakly supervised fine-tuning and model fusion phases. By calculating weighted predictions based on their predicted probabilities, we infer the presence of hallucinations. Table~\ref{tab:vote} presents the details of the voting experiments on the model-agnostic track. Using the same method, we achieve an accuracy of 0.818 on the model-aware track as well. The SFT models are merged models from different training experimental setups, and the detailed training parameters can be found in Appendix~\ref{sec:ap_v4_v7}.

\begin{table}[]
\centering
\begin{tabular}{cc}
\toprule
\textbf{Method} & \textbf{agnostic\_acc} \\
\hline
Mistral-7B-Instruct-v0.2-sft-v4 & 0.810                  \\
Mistral-7B-Instruct-v0.2-sft-v5 & 0.812                  \\
Mistral-7B-Instruct-v0.2-sft-v6 & 0.812                  \\
Mistral-7B-Instruct-v0.2-sft-v7 & 0.814                  \\
voting result       & \textbf{0.834}        \\
\bottomrule
\hline
\end{tabular}
\caption{Model Voting results for the model-agnostic track on the validation set. The voted results outperform any individual model in terms of accuracy.}
\label{tab:vote}
\end{table}

We compared the baseline provided by competition organizers, GPT-4, and our proposed method on the test set in Table~\ref{tab:test}. It is evident that our proposed method outperforms other mehotds, showcasing a significant enhancement in performance.

\begin{table}[h!]
\centering
\begin{tabular}{ccc}
\toprule
\textbf{Method}  & \textbf{agnostic\_acc} & \textbf{aware\_acc} \\
\hline
baseline   & 0.697             & 0.745          \\
GPT-4      & 0.741             & 0.756          \\

our method & \textbf{0.836}    & \textbf{0.805} \\
\bottomrule
\hline
\end{tabular}
\caption{Comparison of methods on the test set.}
\label{tab:test}
\end{table}


\section{Conclusion}
\label{sec:conclusion}
In this paper, we present a unified system for hallucination detection with LLMs when there is no labeled dataset, which wins the 2nd place with an accuracy score of $0.836$ in the model-agnostic track and the 4th place with an accuracy score of $0.8053$ in the model-aware track. To begin with, we generate high-quality weakly-supervised dataset by using large-sized LLMs with prompt engineering and few-shot learning. Then we perform weakly-supervised fine-tuning based on the constructed dataset with different LLMs. Our experiments yield several noteworthy findings:

(1) The quality of the weakly-supervised dataset we construct has a direct impact on the performance of the models in this task. To ensure high-quality training data, we employ multiple large LLMs in the construction process.

(2) Relatively small LLMs can deliver competitive performance in this task when trained on the constructed dataset. However, the performance of small LLMs drops dramatically without fine-tuning.

(3) Using the \textit{MergeKit} tool for model fusion proves to be an effective technique in boosting the performance of hallucination detection.

(4) Employing the model voting method leads to improved performance compared to using a single model alone.

\bibliography{anthology,custom}

\appendix

\section{Instructions in Prompt Engineering}
\label{sec:appendix}

\subsection{Naive version}

\begin{flushleft}
\textit{Context: \{Context\}} \\
\textit{Sentence: \{Sentence\}} \\
\textit{Is the Sentence supported by the Context above? Answer using ONLY yes or no:}
\end{flushleft}

\subsection{Our proposed version}

For the PG task, the instruction is as follows:

\begin{flushleft}
\textit{Given the following information related to Paraphrase Generation task:}\\
\textit{Src: Source input sentence}\\
\textit{Tgt: Paraphrase Generation standard answer}\\
\textit{Hyp: Paraphrase Generation predicted answer} \\
\textit{Please determine whether hyp contains unexpected hallucinations based on src and tgt.} \\
\vspace{12pt}
\textit{Src: \{Src\}} \\
\textit{Tgt: \{Tgt\}} \\ 
\textit{Hyp: \{Hyp\}} \\
\textit{Is the Hyp supported by the Src and Tgt above? Answer using ONLY yes or no:}
\end{flushleft}

For the MT task, the instruction is as follows:

\begin{flushleft}
\textit{Given the following information related to Machine Translation task:}\\
\textit{Src: Source input sentence}\\
\textit{Tgt: Machine Translation standard answer}\\
\textit{Hyp: Machine Translation predicted answer} \\
\textit{Please determine whether hyp contains unexpected hallucinations based on src and tgt.} \\
\vspace{12pt}
\textit{Src: \{Src\}} \\
\textit{Tgt: \{Tgt\}} \\ 
\textit{Hyp: \{Hyp\}} \\
\textit{Is the Hyp supported by the Src and Tgt above? Answer using ONLY yes or no:}
\end{flushleft}

As for the DM task, the instruction is the same as the naive version.

\section{Training Experiment Setup}
\subsection{Constructed dataset}
We constructed a total of \textbf{35,600} weakly supervised samples, ensuring consistency in inference across different LLMs as well as within the same LLM but with different parameter settings.

\subsection{SFT models in Table~\ref{tab:mergekit}}
\label{sec:ap_v1_v3}

The Mistral-7B-Instruct-v0.2-sft-v1, v2, and v3 models are different checkpoint models obtained from the same training setup. These models were trained on a total of 35,600 weakly-supervised data points. The training process utilized a LoRA rank of 32, a learning rate of $3e^{-5}$, and a total of 5 epochs. The training task was executed using 4 A30 GPUs. Specifically, Mistral-7B-Instruct-v0.2-sft-v1, v2, and v3 models were saved at training steps 1000, 3000, and 4000, respectively.

\subsection{SFT models in Table~\ref{tab:vote}}
\label{sec:ap_v4_v7}

The Mistral-7B-Instruct-v0.2-sft-v4, v5, v6, and v7 models are merged models obtained from different training setups. Each model is created by merging two checkpoints from the same setup. 
The v4 model was trained with a LoRA rank of 32, a learning rate of $1e^{-4}$, and a total of 5 epochs. 
The v5 model also had a LoRA rank of 32, a learning rate of $3e^{-5}$, and a total of 5 epochs. 
The v6 model had a higher LoRA rank of 48, a learning rate of $3e^{-5}$, and lasted for 5 epochs. 
Lastly, the v7 model had a LoRA rank of 48, a learning rate of $5e^{-5}$, and a total of 5 epochs. 
All of these models were trained on the constructed dataset.

\section{Lora training VS. Full training}
\label{sec:ap_lora_full}
\begin{table}[h!]
\centering
\begin{tabular}{ccc}
\toprule
\textbf{Method}  & \textbf{agnostic\_acc} & \textbf{aware\_acc} \\
\hline
lora   & 0.806            & 0.790         \\
full     & 0.58            & 0.52          \\
\bottomrule
\hline
\end{tabular}
\caption{Comparison of different training methods based on Mistral-7B-Instruct-v0.2.}
\label{tab:lora_full}
\end{table}

\end{document}